%% file: paper.tex
\documentclass[submission,copyright,creativecommons]{eptcs}

\usepackage[utf8]{inputenc}  
\usepackage[T1]{fontenc}     
\usepackage[english]{babel}  
\usepackage{graphicx}        
\usepackage{cite}            
\usepackage[]{hyperref}            
\usepackage{comment}         
\usepackage{breakurl}        
\usepackage[scaled]{helvet} 
\usepackage{newtxtext,newtxmath} 

\title{Mutation Testing for Industrial Robotic Systems}

\author{%
{Marcela} {Gonçalves dos Santos}, Sylvain Hallé\institute{%
Département d'informatique et de mathématique\\
Université du Québec à Chicoutimi\\
Chicoutimi, Québec, Canada\\
}
\and
F\'abio Petrillo\institute{%
Département de génie logiciel et TI\\
École de technologie supérieure\\
Montréal, Québec, Canada\\
}
}

\def\titlerunning{Mutation Testing for Industrial Robotic Systems}
\def\authorrunning{M.\ Goncalves dos Santos, S.\ Hallé, F.\ Petrillo}
\usepackage{amsmath,amsfonts}

\usepackage{subfig}

\usepackage{algorithm,algpseudocode}
\usepackage{siunitx}
\usepackage{microtype}

\usepackage{xcolor}

\usepackage{multirow}

\newcommand{\lightgrayrow}{}
\newcommand{\silverrow}{}

\newcommand{\ac}[1]{#1}

\usepackage{listings}
\newcommand{\revsym}[1]{%
  \begingroup
  \setlength{\fboxsep}{0.6pt}
  \ifmmode
    \mathchoice
      {\colorbox{red}{$\displaystyle\color{white}{#1}$}}%
      {\colorbox{red}{$\textstyle\color{white}{#1}$}}%
      {\colorbox{red}{$\scriptstyle\color{white}{#1}$}}%
      {\colorbox{red}{$\scriptscriptstyle\color{white}{#1}$}}%
  \else
      \colorbox{black}{\(\color{white}{#1}\)}%
  \fi
  \endgroup
}

\graphicspath{{fig/}}

\begin{document}

\maketitle
\begin{abstract}
Industrial robotic systems (IRS) are increasingly deployed in diverse environments, where failures can result in severe accidents and costly downtime. Ensuring the reliability of the software controlling these systems is therefore critical. Mutation testing, a technique widely used in software engineering, evaluates the effectiveness of test suites by introducing small faults, or mutants, into the code. However, traditional mutation operators are poorly suited to robotic programs, which involve message-based commands and interactions with the physical world. This paper explores the adaptation of mutation testing to IRS by defining domain-specific mutation operators that capture the semantics of robot actions and sensor readings. We propose a methodology for generating meaningful mutants at the level of high-level read and write operations, including movement, gripper actions, and sensor noise injection. An empirical study on a pick-and-place scenario demonstrates that our approach produces more informative mutants and reduces the number of invalid or equivalent cases compared to conventional operators. Results highlight the potential of mutation testing to enhance test suite quality and contribute to safer, more reliable industrial robotic systems.

\end{abstract}


\section{Introduction} 

Industrial robotic systems (IRS) are composed by industrial robots, end-effector (grippers, welding torches, painting heads, cutting tools) and equipment (sensors, vision systems, conveyors). They typically carry out repetitive tasks such as picking and placing, soldering, and cutting. 
A report from the IFR states that installations of new IRS, for the sole year 2022, reached a record-high of 553,052 units, totaling at the time close to 4 million operational units worldwide \cite{IFR2023}.

The increase in the number of industrial robotic systems operating in the most diverse environments also increases the necessity of these systems to handle failures and meet quality aspects. As a matter of fact, robot failures can lead to catastrophic outcomes: 
a study indicates that from 1992 to 2017, at least 41 individuals lost their lives due to industrial robots \cite{LARRY2023}.
Furthermore, according to a study on the reliability of robotic automation, 
industrial robots are non-operational on average 12\% of the time, due to program faults, sensor faults, and other miscellaneous issues \cite{PANAGIOTIS2015}. The cost for \ac{IRS} downtime depends on a variety of factors, but it is estimated that it can cost tens of thousands of dollars in lost production.  Handling failures in \ac{IRS} is therefore crucial for reducing accidents and minimizing the costs associated with lost production.

Engineering software for \ac{IRS} requires compliance with rigorous quality requirements, including reliability \cite{MARIJAN2019}, correctness \cite{BRUGALI2009}, and in particular robustness \cite{GARCIA2020}, defined as the ability to maintain acceptable performance when facing uncertainties, disturbances, or unexpected changes (e.g., sensor noise, environmental variations). In order to ensure satisfaction of these requirements, testing is often considered, as it operationalizes specifications (properties, contracts, or models) as test oracles and finite observations over executions. Testing provides scalable, partial evidence (a form of ``lightweight formal method'') that a system refines or satisfies its intended properties when full verification is infeasible.

Mutation testing is a technique originally developed in software engineering for traditional systems that aims to improve software quality \cite{DBLP:journals/computer/DeMilloLS78}. It involves the systematic introduction of small, syntactically valid changes, called ``mutants'', into a program's source code, simulating potential faults or errors that could occur during program execution. By assessing the effectiveness of a set of tests in detecting and localizing these mutations, mutation testing provides valuable insight into the resilience of a test suite and the overall quality of the software under scrutiny.

Although software developers for traditional systems applied mutation testing and have shown its effectiveness in improving code quality \cite{DBLP:journals/tse/SanchezPSTP24,DBLP:conf/kbse/HeijningenWN0VF24}, its adoption in industrial robotics still needs to be explored. This paper aims to bridge this gap by exploring the applicability and benefits of mutation testing for industrial robotic systems, by proposing an approach going beyond the traditional type of mutants that simply modify the source code. Specifically, we recommend domain-specific mutation operators that capture the semantics of robot actions and sensor readings, therefore generating meaningful mutants at the level of high-level read and write operations, including movement, gripper actions, and sensor noise injection. An empirical evaluation on a simple pick-and-place scenario shows that the proposed operators generate more meaningful mutants while reducing the number of invalid or equivalent cases produced by conventional approaches. These results underline the value of mutation testing as a means to improve test suite effectiveness and ultimately foster safer and more dependable industrial robotic systems.


\section{Background}\label{sec:background} 

In this section, we first define the basics of the two fields that this paper seeks to reconcile, namely industrial robotic systems and their programming, as well as mutation testing techniques taken from the field of software engineering.

\subsection{Industrial Robotic Systems}

\begin{figure}
\centering
 \includegraphics[width=6cm]{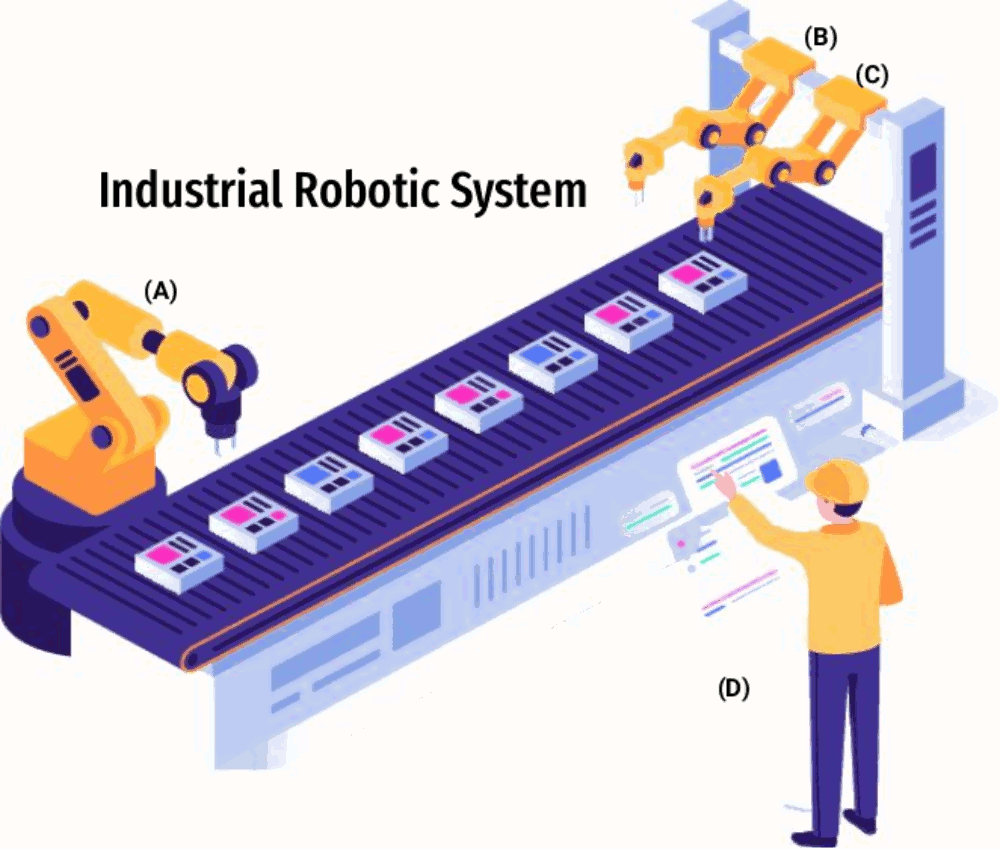}
\caption{The main components of an industrial robotic system.}
\label{fig:IRS}
\end{figure}

An industrial robotic system (IRS) is composed of industrial robots, end-effectors (grippers, magnets, vacuum heads), sensors (visual, torque, collision detection, 3D vision) and equipment (belt conveyors). As with any robot in general, an industrial robot is a complex system composed of both hardware and software components. Figure \ref{fig:IRS} shows a system composed by three industrial robots (A, B and C) that performed tasks to automate industrial processes through a software. Each robot is equipped with a gripper and the whole system also has a conveyor. Moreover, a human (D) interacts with the IRS through a software interface.

Industrial robotic systems (IRS) implement sense–plan–act loops with closed-loop decision-making from sensors to actuators, often in partially structured, time-varying environments (e.g., vision-guided bin picking, conveyor tracking, human–robot collaboration). By the definition of ISO 8373, industrial robots are \emph{automatically controlled, reprogrammable, multipurpose manipulators}, which places them on the autonomy spectrum even when tasks are supervised or constrained. Modern IRS select actions based on perception, handle uncertainty (noise, delays, occlusions), and must guarantee safety under autonomy-like operating conditions—core concerns of autonomous systems research.

Common tasks performed by industrial robots include pick-and-place operations, welding, soldering, painting, palletizing, assembling, and inspection \cite{ISO8373}. These tasks typically involve interactions with the physical environment and must accommodate varying degrees of uncertainty in object positioning, sensor input, and timing.
Among these, pick-and-place is one of the most fundamental and commonly automated tasks, making it especially relevant in production lines, packaging systems, and laboratory automation. This operation is typically divided into four distinct phases \cite{MNYUSIAWALLA2020}: (i)~pre-grasping, in which the gripper is moved to the object's initial position; (ii)~grasping, where the robot securely grips the object, breaking its contact with the original surface; (iii)~transport, during which the object is moved to the destination; and (iv)~placement, where the robot releases the object at the designated location. Each of these phases requires precise coordination of sensing, motion control, and timing, making pick-and-place a representative and meaningful task for evaluating robotic system performance.

\subsubsection{Hardware Components}

The mechanical structure of a robot, which includes joints, links, and the end effector, serves as the foundation for movement and task execution. In typical robotics applications, the position of the end effector (such as a gripper or tool) within the workspace, along with the orientation and configuration of the joints, is used to describe the robot's state. 


Sensors monitor the robot's state, including joint positions, velocities, and torques. Common examples are encoders and potentiometers, which measure joint angles to ensure precise movement and positioning. The integration of sensors is crucial for allowing robots to perceive and interact with their surroundings. However, the accuracy and reliability of sensor data can deteriorate over time or fluctuate based on environmental conditions and manufacturing tolerances \cite{HOSSAIN2024}. Numerous case studies illustrate the critical relationship between sensor precision and calibration in robotics \cite{LUBRANO2011, ICLI2020, LI2021}.

In addition to sensor inaccuracies, it is important to consider the concept of an error threshold in robotic applications. The error threshold determines the maximum allowable deviation between the robot's actual position and the position specified by the task requirements \cite{MORDECHAI2018}. For instance, if a robot arm is programmed to place an object at a specific location, the robot's end effector must be within a close interval from the target for the task to be considered successful. If the deviation surpasses this threshold, the task is classified as unsuccessful, necessitating corrective actions.

\subsubsection{Software Components}

The scope of our study is on the software component, which plays a critical role in controlling robotic movements, processing data from sensors, and ensuring the precise execution of tasks. This component is itself composed of two parts. First, the \emph{control layer} is responsible for translating commands so that the IRS can understand and execute them; thus, it is essentially a collection of drivers interacting with the hardware. Second, the \emph{application layer} is composed of a software script, which defines the robot's desired behavior according to business requirements \cite{AHMAD2016, ISO8373}.

\begin{figure}
\centering
\includegraphics[width=\linewidth]{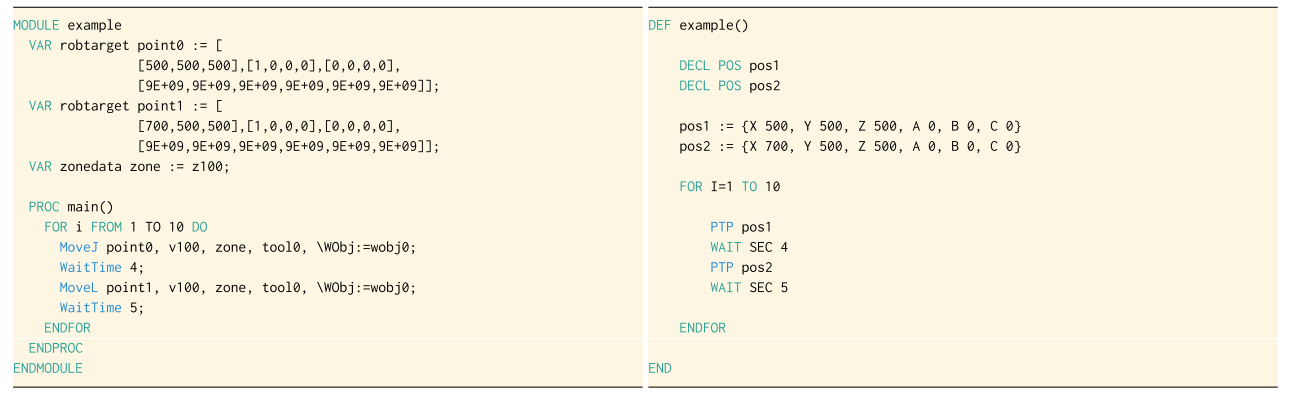}
\caption{Examples of programs written in two IRPLs: ABB's RAPID (left) and KUKA's KRL (right) \cite{POGLIANI2020}.}
\label{fig:examplIRPL}
\end{figure}

The latter layer is typically expressed in a language specialized for robot control, and includes instructions for moving the robot's arm(s) and accessing low-level resources \cite{POGLIANI2020}. Figure \ref{fig:examplIRPL} presents two examples of industrial robot programming utilizing ABB's RAPID and KUKA's KRL. While both ABB's RAPID and KUKA's KRL are designed for similar functions (namely, controlling robotic movements and operations), they differ significantly in syntax and structure. 

Another way of programming an IRS is through the use of the Robotic Operating System (\ac{ROS}) \cite{ROSBOOK}, a standard framework for developing robotic software. A system created with this framework is made from different programs running simultaneously and communicating with one another through message passing. \ac{ROS} therefore acts as a middleware between robot program and the robot hardware. This is the framework we shall concentrate on in the remainder of this paper.

\subsubsection{Simulation}

Simulators 
play an important role in the development and validation of robotic behaviors prior to deployment. 
Timperley \textit{et al.\@} contend that many real-world robotics bugs could be replicated and addressed in simulation environments \cite{TIMPERLEY2018}. Moreover, simulations mitigate the risk of damaging equipment \cite{JAMES2023}, eliminate the necessity for physical prototypes \cite{ROTH2023}, and offer a cost-effective means to implement changes \cite{ROBERT2020}. In the context of the present paper, the use of simulators is all the more important, since our goal is to apply the mutation testing methodology to IRS software. As we already mentioned, mutation testing involves running a large number of variants of a program, which we consider impractical to conduct in the physical world.

\begin{figure}
\centering
\includegraphics[width=4in]{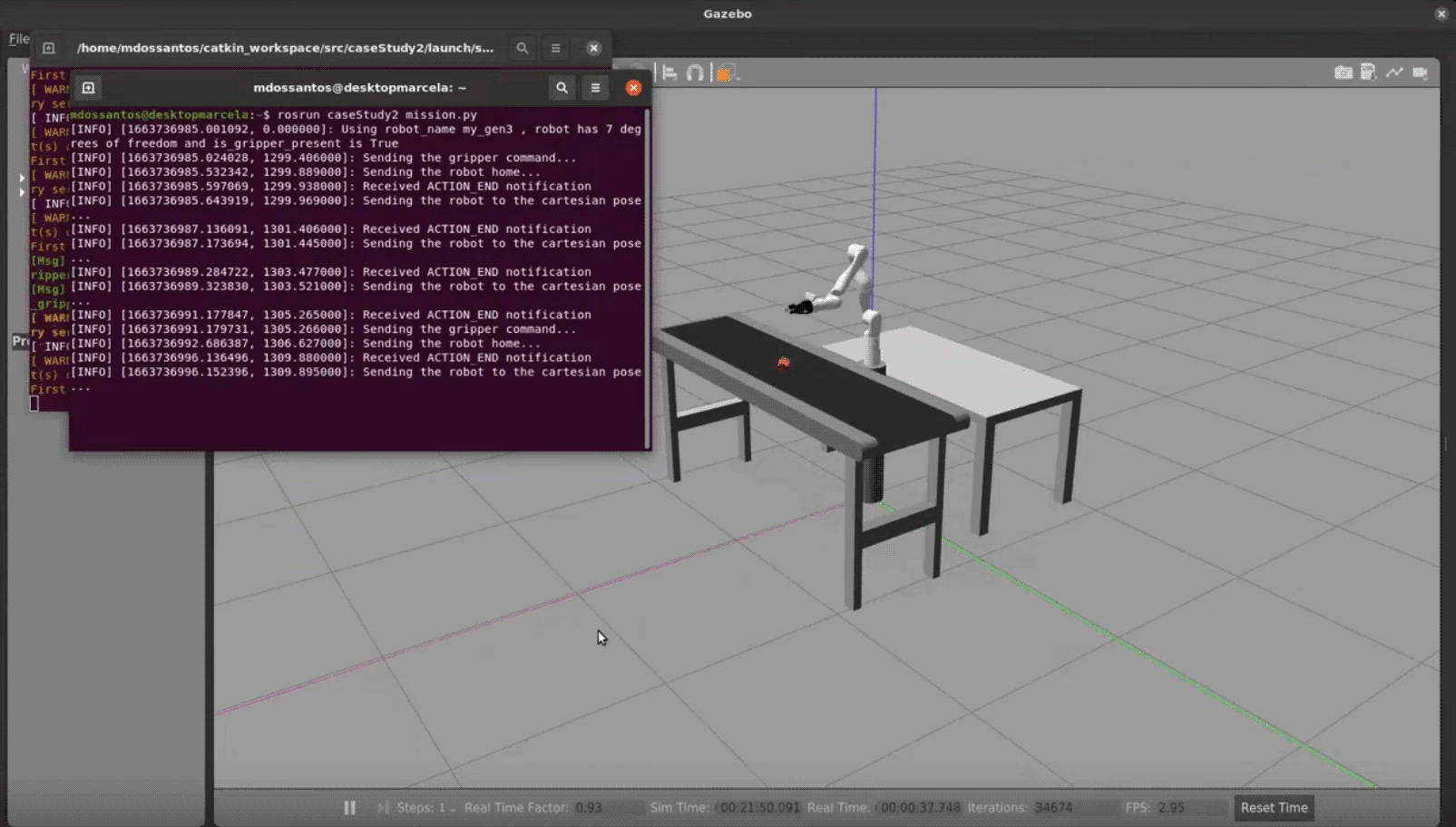}
\caption{A screenshot of ROS and Gazebo simulating the operation of a Kinova Gen3 gripper arm.}
\label{fig:screenshot}
\end{figure}

There exist multiple simulation platforms \cite{michel2004,rohmer2013,todorov2012}; among the available options, Gazebo emerged as the leading choice for robotic simulation due to its ability to replicate diverse robotic platforms equipped with standard sensors like cameras, GPS units, and IMUs \cite{gazebo}. Despite operating independently, Gazebo seamlessly integrates with ROS via a software package enabling bidirectional communication \cite{ROSBOOK,FARLEY2022}. Figure \ref{fig:screenshot} shows a typical setup: ROS commands are sent to the simulator through a terminal, while the main interface shows a real-time rendition of the simulated scenario, in the present case a gripper arm and a colored box.\footnote{A demonstration of this scenario is available online: \url{https://www.youtube.com/watch?v=9S0R0mGKA-I}}

\subsection{Mutation Testing}

Software testing involves systematically executing a piece of software under controlled conditions to observe its behavior, identify faults, and assess its robustness. Testing plays an important role in improving confidence in a program's stability and fault tolerance \cite{SQQA2018}.

Mutation testing is the particular technique that consists of transforming a program under test into several slightly altered versions (the so-called ``mutants''), and determining whether the test suite developed for the program is able to distinguish these modified versions from the original program \cite{DBLP:journals/computer/DeMilloLS78}. A mutant program is said to be ``killed'' if at least one of the test cases fails on this program, illustrating the test suite's ability to detect the modification. Otherwise, the mutant is deemed to have survived the test suite.

\subsubsection{Mutation Operators}

Modifications to a program are applied through a variety of simple functions called \emph{mutation operators}. For example, Figure \ref{subfig:original} shows a simple pseudocode snippet for a function calculating the maximum of two numbers. Figure \ref{subfig:mutant} shows a modified version of this function where a single symbol has been modified: the condition $a>b$ has been inverted to $a<b$ (highlighted in red). The flip from $>$ to $<$ at one location in the code is an example of a mutation operator. Typical modifications include changing a mathematical operator, swapping the order of two statements, or replacing a condition with its negation. Table \ref{tab:operators} gives a summary of the most commonly used program-level mutation operators.

\begin{figure}
\centering
\subfloat[Original program]{%
\begin{minipage}{2.25in}
\begin{algorithmic}[1]
\Procedure{max}{$a,b$}
\If{$a > b$}
  \State \textbf{return} $a$
\Else
  \State \textbf{return} $b$
\EndIf
\EndProcedure
\end{algorithmic}
\end{minipage}\label{subfig:original}
}
\subfloat[Program mutant]{%
\begin{minipage}{2.25in}
\begin{algorithmic}[1]
\Procedure{max}{$a,b$}
\If{$a~\revsym{<}~b$}
  \State \textbf{return} $a$
\Else
  \State \textbf{return} $b$
\EndIf
\EndProcedure
\end{algorithmic}
\end{minipage}\label{subfig:mutant}
}
\caption{A simple program and a possible mutant.}
\label{fig:programs}
\end{figure}

\begin{table}
\centering
\scalebox{0.8}{%
\input operators.inc
}
\caption{The defining set of program-level mutation operators (adapted from \cite{ist}).}
\label{tab:operators}
\end{table}

\emph{Mutation score} is defined as the proportion of mutants killed by the test suite, among all the mutants considered in the process. A high mutation score indicates that the test suite adequately circumscribes the expected behavior of the program. It has indeed been observed empirically that mutants are a good representation of real faults, both in random and coverage-based test suites \cite{DBLP:journals/tse/AndrewsBLN06,DBLP:journals/tse/JiaH11}. While there is no universally accepted threshold, the software testing literature typically considers a mutation score exceeding 80\% as indicative of good test quality, with scores above 90\% reflecting exceptional fault detection capabilities \cite{PAPADAKIS2018, DBLP:journals/tse/JiaH11,DBLP:conf/issta/UntchOH93}. 

\subsubsection{Effective Mutation Operators}

Performance remains the primary limitation of mutation testing, with long execution times hindering agile processes \cite{icsoft-ea14,DBLP:journals/tse/SanchezPSTP24}. The exhaustive application of mutation operators on every applicable code point generates a prohibitively large number of mutants for any nontrivial program. Therefore, a large amount of work in the field concentrates on the reduction of the number of mutants, while still preserving the fault-detection capability of the remaining instances \cite{icsoft-ea14,MA201618}. Some approaches consider a random or constrained selection of mutants \cite{acree79,DBLP:conf/sbes/WongMDM94}, or the introduction of more than one fault per mutant \cite{DBLP:conf/icst/HarmanJL10}.

A particular reduction strategy is to act at the level of mutation operators, and select a subset of so-called \emph{effective} operators \cite{DBLP:journals/tosem/OffuttLRUZ96}. The goal is to restrict mutation operators to only those that are representative of likely faults in the program to test, a property that is obviously specific to the domain of application. As a simple example, mutation operators producing syntactically incorrect pieces of code (such as replacing a string value by a number, or \textit{vice versa}) may be appropriate when testing a compiler (which should be able to reject these incorrect programs), but make little sense when testing the logic of the program itself.


\section{Mutation Testing Applied to IRS}\label{sec:contribution} 

Conventional software systems are typically designed to operate in discrete, well-defined digital environments. Inputs, outputs, and system states are predictable and can often be specified with precision. However, the straightforward application of Software Engineering principles to IRS programming presents numerous challenges.

By definition, an \ac{IRS} is an integration of hardware, software with the physical components that must operate in the physical world. The software in a robot needs to interact with a world that is continuous, dynamic, and full of uncertainty \cite{GROOVER2007}. Furthermore, robots sense and act upon the physical world using components like cameras, force sensors, and motors—all of which are inherently noisy and imprecise. Unlike a web or mobile app, a robotic arm may misplace an object due to sensor noise or slight mechanical drift. These systems are also sensitive to timing: a delay of a few milliseconds in a control loop can cause a robot to miss or mishandle a task \cite{LI2016}.
In addition, the state space in which robots operate (that is, the real world) is vast and can not be exhaustively tested or simulated. As a result, emergent behaviors (unexpected actions that arise from complex interactions between components) are common and often difficult to predict or replicate \cite{AFZAL2020}.

We have seen that mutation testing is a well-established technique in software engineering for assessing the strength of test suites by introducing artificial faults. When applied to industrial robotic systems (IRS), however, conventional mutation operators face limitations due to the specific structure of robotic programs and their close interaction with the physical environment. This section examines how mutation testing can be adapted to IRS and outlines the challenges that arise when using standard operators in this domain.

\subsection{Limitations of Classical Mutation Testing}\label{subsec:limitations}

\begin{figure}
\centering
\includegraphics[width=5in]{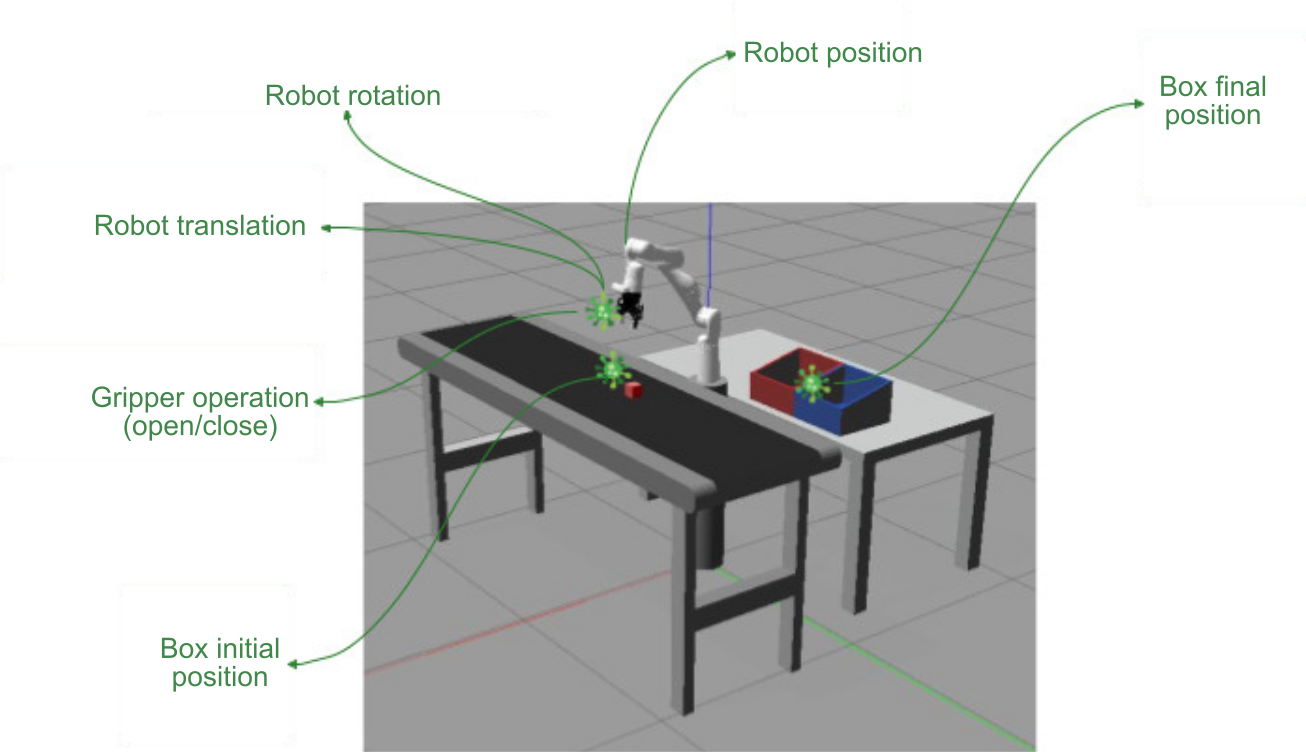}
\caption{The pick-and-place scenario used in our experiment, with the localization of points where mutation operators are applied.}
\label{fig:robot_mutants}
\end{figure}

Consider a scenario where a robot equipped with a moveable grasp whose task is to pick a box and place it on one of two scales placed to its left and its right, depending on its color. Figure \ref{fig:robot_mutants} illustrates this scenario, while Figure \ref{fig:simple-program} shows a pseudocode of a possible program for this robot, which uses a message-based interaction mechanism in the style of the Robot Operating System (ROS) \cite{ROSBOOK}. The robot picks the box, lifts it from the ground by some amount, and does the reverse operations after turning to two possible angles, depending on the box's color determined by some visual sensor.

\begin{figure}
\centering
\begin{minipage}{2.25in}
\begin{algorithmic}[1]
\State \textsl{send}(\texttt{``pick''})
\State \textsl{send}(\texttt{``lift/5''})
\If{\textsl{read}(\texttt{``color''}) = \texttt{``red''}}
  \State \textsl{send}(\texttt{``turn/90''})
\Else
  \State \textsl{send}(\texttt{``turn/270''})
\EndIf
\State \textsl{send}(\texttt{``drop/5''})
\State \textsl{send}(\texttt{``release''})
\end{algorithmic}
\end{minipage}
\caption{Pseudocode for a simple robot program.}
\label{fig:simple-program}
\end{figure}

In this context, a test \emph{case} consists of an initial environment (location of boxes and initial state of the robot), as well as a set of conditions. These conditions must be satisfied either at the end of the test, or optionally at any other moment during the execution. A test \emph{suite}, as usual, is understood as a set of test cases.
As with any other type of program, it may be desirable to perform mutation testing on the test suite to assess its effectiveness. A straightforward way of doing so is to consider the listing of Figure \ref{fig:simple-program} as an arbitrary piece of procedural code, and to apply the commonly defined mutation operators to generate mutant versions of the original program. Multiple issues can be identified in such a case.

A first limitation of applying traditional mutation operators to robotic programs is that there are very few points in the code where such mutations can be injected. These programs often consist of simple, sequential instructions such as sending commands to the robot or reading values from sensors. Unlike conventional software, they make limited use of variables, arithmetic expressions, or complex control structures, which are the usual targets of mutation. As a result, the opportunities for introducing meaningful mutations are much more constrained.

A second issue is that much of the program logic is embedded inside string parameters representing robot commands. If mutation operators are applied naively at the character level, they tend to generate syntactically invalid instructions --for example, altering the command “pick” into “pic.” Such mutants are immediately rejected by the robot or simulator, causing every test case to fail without providing any useful feedback about the adequacy of the test suite. This produces noise rather than insight.

Third, even when a mutation results in a syntactically valid command, the altered behavior may be physically impossible to execute. For instance, modifying the lifting height of a gripper from 5 units to 6 could exceed the robot's operational range, or lowering a box by an extra unit might imply placing it below the floor level. These cases do not expose weaknesses in the test suite; instead, they highlight physical infeasibility, which again reduces the value of such mutants.

Finally, the opposite situation can also occur: a mutation may be valid but fail to introduce any meaningful behavioral difference. For example, replacing a rotation of 90 degrees with 91 degrees may still cause the robot to place an object in the correct location, producing results indistinguishable from the original program. Such equivalent mutants inflate the number of generated cases without improving fault detection, thereby wasting computational effort and obscuring the effectiveness of the mutation analysis.

In this example, a direct application of mutation operators for arbitrary procedural code would therefore produce a suite of mutant programs such that a large proportion of them are either guaranteed to be killed, or are equivalent to the original program. In either case, these mutants provide no added value as they are unable to reveal any new gap in the coverage achieved by the test suite. This has long been identified as a problem that reduces the effectiveness of mutation testing \cite{DBLP:conf/issta/UntchOH93}, and is exacerbated by the fact that tests on robotic systems are usually lengthy and may require heavy preparations before each run (if executed on actual physical components).

\subsection{Domain-specific Mutation Operators}

Better yet would be to define mutation operators taking into account the fact that the listing of Figure \ref{fig:simple-program} is not just any piece of procedural code, but a specific type of program that manipulates robotic components interacting with the physical world. In this context, one could understand that the numbers \texttt{90} and \texttt{270} mean ``left'' and ``right'', and that confusing one for the other is probably more likely for a developer than providing an angular value that is incorrect by a single degree.\footnote{Although in this specific example, negating the condition of line 3 would result in the same observable behavior.}

The first step in adapting mutation testing to industrial robotic systems is to identify the set of high-level write operations available to the robot, that is, the commands that actively modify the state of the simulation or the physical system. In a pick-and-place scenario, such operations include moving the arm to a predefined location, translating it by a given offset, rotating it to face the delivery table, or operating the gripper to grasp and release a box. Complementary to these are the read operations, which probe the state of the system and provide information about the robot's environment or its own configuration. For the pick-and-place task, examples include reading the current position of the box on the conveyor belt, detecting its color to decide which target location to use, or measuring the angular orientation of the robot arm before initiating transport. These observations guide the control logic that determines the sequence of movements to perform.

Once these categories of operations are established, the next step is to define mutation operators that meaningfully perturb them. For write operations, this may involve inverting a movement direction (e.g., rotating right instead of left), suppressing an action (e.g., skipping the gripper-closing command), or repeating an action (e.g., attempting to release the box twice). For read operations, mutations can be introduced by returning an incorrect color classification, injecting Gaussian noise into position readings, or flipping the sign of a measured distance. In each case, the goal is to create mutants that reflect plausible faults encountered in robotic programming or sensor interactions, while avoiding trivial or nonsensical behaviors. Table \ref{tab:mutantsoperations} summarizes the domain-specific mutation operators we have identified.

\begin{table}
\centering
\input irs-operators.inc
\caption{Mutation operators specific to robotic systems.}
\label{tab:mutantsoperations}
\end{table}


\section{Implementation and Experiments} 

The use of industrial robots in pick-and-place scenarios is a common occurrence in competitions and benchmarking exercises. Our decision to draw inspiration from a robotic competition is rooted in the limited accessibility of real-world industrial requirements, as noted by \cite{TIMPERLEY2018}. In this section, we describe and discuss the experimental results obtained.

\subsection{Experimental Setup}

For direction, we looked into the Robotic Grasping and Manipulation Competition's Task Pool. In particular, we turned to a specific task outlined in their competition framework: \textit{Pick Up and Place Using Tongs} \cite{SUN2018}, whose outline is similar to the picking task already discussed in Section \ref{subsec:limitations}.

The goal is to ensure that the box with a specific color on the conveyor belt is dropped at the delivery point, respecting the box color. The robot needs to be positioned correctly before the pick-and-place process can begin within a threshold of 0.02 cm in all three axes. For the box the threshold is 0.01 cm also for the three axes. The final position of the box should not exceed 0.02 cm in any of the three~axes.

The robot model used was the Gen3 from Kinova (\cite{KINOVA}), and the end-effector is a gripper, the Robotiq-2f-85 from (Robotiq \cite{ROBOTIQ}). This model has the following sensors: torque, position, current, voltage, temperature, accelerometer and gyroscope. We use the position sensors that give to us the arm position for the three axes. To communicate with the Kinova Gen3 robot, we use the \textit{Kinova Kortex} API integrated with ROS through the package \textit{ROS Kortex}. Thus, ROS Kortex allows us to use the ROS mechanisms (communication and file system) and Gazebo as a simulator to perform the experiments. The source code of the experiments is available online\footnote{\url{https://github.com/mgdossantos/aat4irs_v3}}.

To evaluate our approach, we implemented two families of mutation operators: the conventional operators typically applied to procedural code, and our proposed domain-specific operators tailored for industrial robotic systems. Both were applied to a simple pick-and-place program executed by a simulated robot arm, using the same test suite to ensure comparability. This setup allowed us to contrast the nature, volume, and effectiveness of the mutants produced by each approach.

\subsection{Results}

The first dimension of comparison was the number and type of mutants generated. Conventional operators produced a large pool, but many of these were either syntactically invalid or semantically meaningless in the context of robotic execution. By contrast, our domain-specific operators generated a more compact yet relevant set of 26 mutants, as summarized in Table \ref{tab:mutants}. These mutants covered a broad range of operations, including translation and rotation changes, gripper manipulations, and modifications to both initial and final positions of the robot and the box. This is to be contrasted with ``classical'' mutation operators (cf.\ Table \ref{tab:operators}), which would have resulted in the generation of countless mutants for each of the string parameters --most of them being flatly invalid at the onset.

We then examined the proportion of mutants killed by the test suite. Because the domain-specific operators were designed with the semantics of robot actions and sensor behaviors in mind, they yielded mutants that more accurately reflected realistic faults. As a result, they provided a more stringent challenge for the test cases. For instance, modifying rotation directions, altering gripper behavior, or introducing Gaussian noise into position readings often led to failures that exposed weaknesses in the test suite.

\begin{figure}
\centering
\includegraphics[width=.375\linewidth]{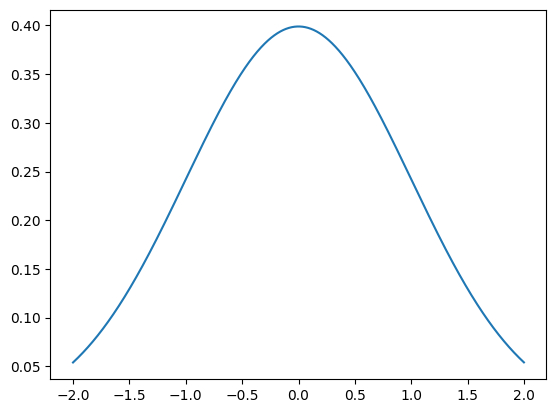}
\caption{Probability distribution of the noise added to the sensor readings.}
\label{fig:noise}
\end{figure}

For the noise added to sensor readings, we simulated Gaussian noise, which is normally distributed and commonly used to model measurement errors or communication noise. The distribution we applied is shown in Figure \ref{fig:noise}, where the lowest values are around 0.053 and the highest reach approximately 0.39. This form of perturbation ensured that the mutants were not only syntactically valid but also representative of realistic disturbances in robotic sensing.

In order to guarantee both the randomness of the injected noise and the diversity of box placements, we executed the test suite for each of the 26 mutants five times. The results of these runs, summarized in Table \ref{tab:mutantscoreperround}, indicate mutation scores ranging from 77\% to 81\%. This repetition confirmed the stability of the results while also demonstrating the discriminating power of the generated mutants, which were neither trivially killed nor trivially equivalent.

The evaluation also considered execution feasibility. Mutants produced with conventional operators frequently led to physically impossible scenarios, such as placing a box below ground level or exceeding the robot's movement range. In contrast, our operators explicitly accounted for robotic constraints, as illustrated in Figure \ref{fig:robot_mutants}, which shows the specific points of localization and operations where mutations were applied. This ensured that most mutants could be executed and provided meaningful feedback rather than producing uninformative crashes.

Overall, the experiments show that tailoring mutation operators to the semantics of industrial robotics significantly improves the relevance of generated mutants and makes the mutation score a more reliable indicator of test suite adequacy. By capturing faults that closely mirror realistic errors, domain-specific mutation testing offers a practical means to strengthen the dependability of industrial robotic systems.

\begin{table}
\centering
\scalebox{0.8}{%
\input mutants.inc
}
\caption{The list of mutants generated in our experiment.}
\label{tab:mutants}
\end{table}

\begin{table}
\centering
\begin{tabular}{ccc}
\hline
\textbf{Round}&\textbf{Score}\\
\hline
\lightgrayrow
\#1 & 92\%\\
\hline
\silverrow
\#2 & 85\%\\
\hline
\lightgrayrow
\#3 & 92\% \\
\hline
\silverrow
\#4 & 88\% \\
\hline
\lightgrayrow
\#5 & 88\% \\
\hline
\end{tabular}
\caption{Mutation score for each round.}
\label{tab:mutantscoreperround}
\end{table}

\subsection{Threats to Validity}

\subsubsection{Internal Validity}

Internal validity concerns whether observed effects are genuinely caused by the intervention rather than by confounders (e.g., selection bias, learning effects, instrumentation); typical mitigations include randomization, controls/baselines, and consistent procedures.

Our results depend heavily on the fidelity of the Gazebo simulator. While Gazebo provides realistic physics and sensor modelling, it cannot perfectly capture all uncertainties and constraints of real-world environments, such as hardware wear, calibration drift, or unpredictable environmental disturbances. Thus, the real-world applicability of the proposed mutation operators remains an open question until they are validated on physical robotic platforms.

\subsubsection{External Validity}

External validity addresses generalizability—whether findings are likely to hold for other projects, teams, technologies, datasets, or environments—supported by careful sampling, context description, and replication across settings.

The mutation operators proposed in this study were specifically tailored for a pick-and-place task. Although this task is representative and widely used in industrial robotics, other domains such as welding, painting, or collaborative human-robot interaction involve different semantics, constraints, and error patterns. Therefore, the generalizability of the operators to these scenarios has not been demonstrated and requires further investigation.

\subsubsection{Construct validity}

Finally, construct validity asks whether the operational measures truly capture the intended theoretical concepts.

Our evaluation primarily relies on the mutation score as a measure of test effectiveness. While mutation score is a well-established metric in software testing, we did not investigate its correlation with the detection of real-world faults in robotic systems, whether through seeded bugs or historical incident data. As a result, it is uncertain to what extent improvements in mutation score translate into actual fault detection capability in industrial robotic contexts.


\section{Related Work} 

An extensive survey by Jia and Harman \cite{DBLP:journals/tse/JiaH11} reports that mutation testing has been ported, as a white-box testing technique, to most modern procedural languages. As a matter of fact, defining mutation operators specific to a particular application domain has already been suggested in a variety of fields, such as event processing queries \cite{DBLP:conf/issre/Gutierrez-MadronalSZDM12}, the Google Query Language \cite{DBLP:conf/aciids/Gutierrez-Madronal20}, web service compositions expressed in WS-BPEL \cite{DBLP:conf/ecows/Dominguez-JimenezEGM09}, software product lines \cite{DBLP:journals/stvr/KrugerALS19}, and models for the SMV model checker \cite{DBLP:conf/icfem/AmmannBM98}.

Curiously, and despite the issues raised in Section \ref{subsec:limitations}, very few works have addressed the problem of devising appropriate mutation operators specific to robot programming languages like ROS. We are aware of only two such approaches. The first is a mutation testing framework for \textit{RoboChart} \cite{Hierons2021}. It seeds a fault into specification $S$ and uses the Wodel tool \cite{GOMEZABAJO2025102195} to generate model mutants. It then translates $S$ and each mutant $S'$ to CSP with RoboTool, and finally uses the FDR model checker in order to check whether $S$ and $S'$ are equivalent. If not, FDR returns a counterexample trace that directly becomes a test case that can kill $S'$. This results in an automated pipeline with guaranteed detection of the seeded faults and the ability to discard “refining/equivalent” mutants.

In spirit, our work is similar in using mutation to assess and strengthen tests for robotic systems; however, the location and nature of mutations differ: whereas \cite{Hierons2021} mutate \emph{models} (state-machine level) and rely on formal semantics and refinement to synthesize tests, our work mutates \emph{program-level interactions} specific to industrial robotic systems, defining domain-aware operators over high-level write/read actions.

On its side, ROSMutation presents a ROS-specific mutation testing library integrated into a fault-injection tool \cite{rosmutation}. It scans ROS-Py code to find lines that use ROS APIs (publishers, subscribers, services, parameters, time, logs) and generates logical mutants (e.g., topic/parameter substitutions, queue-size/time edits). Two evaluations are reported: (i) 25 tutorial ROS scripts (1,186 mutants; most survived, few killed) and (ii) the SRVT package (1,260 ROS mutants; $\approx$41\% mutation score).

One can observe at the outset that the mutation score is relatively low compared to our (arguably preliminary) experimental evaluation. This could be explained by the fact that ROSMutation operates at the ROS API/code level to gauge robustness, while our contribution targets execution semantics of industrial robotic systems test suite adequacy. Concretely, we define language-agnostic operators over high-level write/read actions designed to avoid trivial or invalid mutants under physical constraints. In contrast, ROSMutation focuses on API-aware edits that ensure syntactic plausibility but can yield many survivors reflective of ROS fault tolerance. 


\section{Conclusion}\label{sec:conclusion} 

Traditional mutation operators perform poorly on robotic programs because they ignore the program's tight coupling with sensors, actuators, and physical constraints. We addressed this gap by defining domain-specific operators for industrial robotic systems that act on high-level write/read actions (movement, rotation, grasp/release, and sensor observations). In a pick-and-place scenario, these operators yielded informative mutants while sharply reducing invalid and equivalent cases, thereby improving the diagnostic power of mutation testing for robotics. However, we acknowledge that defining a complete set of domain-specific mutation operators could be challenging, as this process involves numerous design choices.

Future work will focus on a dedicated mutation toolchain for robots and the physical world, integrating simulators and hardware-in-the-loop to automate mutant generation, execution, and analysis. We will broaden the evaluation to more complex tasks (e.g., cluttered bin picking, multistep assembly, multi-robot coordination) and richer sensing/actuation modalities. Finally, we propose \emph{physical mutants} for real-world testing—controlled perturbations such as slight fixture misalignments, payload mass/CoM shifts, belt-speed jitter, calibrated sensor bias or dropout, occlusions, and bounded actuation delays—executed under safety guards. Such physically grounded mutations can expose gaps that purely software-level operators miss, moving mutation testing for robotics closer to the realities of deployment.





\end{document}

%% file: operators.inc
\begin{tabular}{|l|p{9.5cm}|p{6.5cm}|}
\hline
\textbf{Operator} & \textbf{Description} & \textbf{Example (before $\to$ after)} \\
\hline
ABS & Insert absolute value functions: \texttt{abs()}, \texttt{negAbs()}, \texttt{failOnZero()} & \texttt{x = 3 * a;} $\to$ \texttt{x = abs(3 * a);} \\
\hline
AOR & Replace arithmetic operators (+, –, *, /, \%, **) with each other or with \texttt{leftOp}, \texttt{rightOp}, \texttt{mod} & \texttt{x = a + b;} $\to$ \texttt{x = a - b;} \\
\hline
ROR & Replace relational operators (<, $\leq$, >, $\geq$, ==, $\neq$) with each other, \texttt{falseOp}, or \texttt{trueOp} & \texttt{if (m > n)} $\to$ \texttt{if (m <= n)} \\
\hline
COR & Replace logical operators (\texttt{\&\&}, \texttt{||}, \texttt{\&}, \texttt{|}, \texttt{\^{}}) with each other, \texttt{falseOp}, \texttt{trueOp}, \texttt{leftOp}, \texttt{rightOp} & \texttt{if (a \&\& b)} $\to$ \texttt{if (a || b)} \\
\hline
SOR & Replace shift operators (\texttt{<<}, \texttt{>>}, \texttt{>>>}) with each other or with \texttt{leftOp} & \texttt{x = m << a;} $\to$ \texttt{x = m >> a;} \\
\hline
LOR & Replace bitwise operators (\texttt{\&}, \texttt{|}, \texttt{\^{}}) with each other, \texttt{leftOp}, or \texttt{rightOp} & \texttt{x = m \& n;} $\to$ \texttt{x = m | n;} \\
\hline
ASR & Replace assignment operators (=, +=, -=, *=, /=, \%=, \&=, |=, \^{ }=, <<=, >>=, >>>=) with each other & \texttt{x += 3;} $\to$ \texttt{x -= 3;} \\
\hline
UOI & Insert unary operators (+, –, !, $\sim$) before expressions & \texttt{x = 3 * a;} $\to$ \texttt{x = -(3 * a);} \\
\hline
UOD & Delete unary operators (+, –, !, $\sim$) & \texttt{if (!(a > -b))} $\to$ \texttt{if (a > -b)} \\
\hline
SVR & Replace each variable reference with another variable of the same type in scope & \texttt{x = a * b;} $\to$ \texttt{x = c * b;} \\
\hline
BSR & Replace a statement with \texttt{Bomb()} to force execution & \texttt{x = a * b;} $\to$ \texttt{Bomb();} \\
\hline
\end{tabular}

%% file: irs-operators.inc
\begin{tabular}{ p{4.5cm}  p{7.5cm}} 
\hline
\textbf{Original }&\textbf{Mutant}\\\hline
\lightgrayrow
rotate left by $\theta$ & rotate \textbf{right} by $\theta$\\\hline
\silverrow
rotate right by $\theta$ & rotate \textbf{left} by $\theta$ \\\hline
\lightgrayrow
translate forward by $\delta$ & translate \textbf{backwards} by $\delta$ \\\hline
\silverrow
translate backwards by $\delta$ & translate \textbf{forward} by $\delta$ \\\hline
\lightgrayrow
translate forward by $\delta$ & translate forward \textbf{to} $\delta$ \\\hline
\silverrow
translate backwards by $\delta$ & translate backwards \textbf{to} $\delta$ \\\hline
\lightgrayrow
do command $x$ & do \textbf{nothing} \\\hline
\silverrow
do command $x$ & do command $x$ \textbf{twice} \\\hline
\lightgrayrow
distance value is $x$ & distance value is $-x$ (reverse direction) \\\hline 
\silverrow
distance value is $x$ & distance value is $x$ + noise \\\hline
\end{tabular}

%% file: mutants.inc
\begin{tabular}{ccl}
\hline
\textbf{Mutant} & \textbf{Type} & \textbf{Description} \\
\hline
1  & Translation & Change the $y$-value in translation \\
\hline
2  & Rotation    & Change the angle orientation in rotation \\
\hline
3  & Translation & Change the $z$-value in translation \\
\hline
4  & \multirow{3}{*}{Gripper Operation} & Do not change gripper status \\
5  &                                        & Change gripper status twice \\
6  &                                        & Change gripper status with the opposite expected operation \\
\hline
7  & Rotation    & Change angle orientation in rotation \\
\hline
8  & Translation & Change $x$-value in translation \\
\hline
9  & \multirow{6}{*}{Robot Initial Position} & Sensor reading with opposite expected value for $x$-component \\
10 &                                           & Sensor reading with opposite expected value for $y$-component \\
11 &                                           & Sensor reading with opposite expected value for $z$-component \\
12 &                                           & Sensor reading with noise in $x$-component \\
13 &                                           & Sensor reading with noise in $y$-component \\
14 &                                           & Sensor reading with noise in $z$-component \\
\hline
15 & \multirow{6}{*}{Box Initial Position} & Sensor reading with opposite expected value for $x$-component \\
16 &                                        & Sensor reading with opposite expected value for $y$-component \\
17 &                                        & Sensor reading with opposite expected value for $z$-component \\
18 &                                        & Sensor reading with noise in $x$-component \\
19 &                                        & Sensor reading with noise in $y$-component \\
20 &                                        & Sensor reading with noise in $z$-component \\
\hline
21 & \multirow{6}{*}{Box Final Position} & Sensor reading with opposite expected value for $x$-component \\
22 &                                       & Sensor reading with opposite expected value for $y$-component \\
23 &                                       & Sensor reading with opposite expected value for $x$-component \\
24 &                                       & Sensor reading with noise in $x$-component \\
25 &                                       & Sensor reading with noise in $y$-component \\
26 &                                       & Sensor reading with noise in $z$-component \\
\hline
\end{tabular}